\documentclass[letterpaper, 10 pt, conference]{ieeeconf}  

\IEEEoverridecommandlockouts                              

\usepackage{graphics} 
\usepackage{epsfig} 
\usepackage{multirow}
\usepackage{caption}
\usepackage{subcaption}
\usepackage{hyperref}
\usepackage{multicol}
\usepackage[]{todonotes}

\newcommand{\Janik}[1]{\todo[inline,color=orange!70]{Janik: #1}}
\renewcommand{\Janik}[1]{#1}

\title{\LARGE \bf
GraspME - Grasp Manifold Estimator
}

\author{Janik Hager$^{1}$, Ruben Bauer$^{1}$, Marc Toussaint$^{2,3}$, Jim Mainprice$^{1,2}$\\
\authorblockA{$^1$Machine Learning and Robotics Lab, IPVS, University of Stuttgart, Germany}
\authorblockA{$^2$Max Planck Institute for Intelligent Systems ;  IS-MPI ; T{\"u}bingen/Stuttgart, Germany}
\authorblockA{$^3$Technische Universit\"at Berlin ; TUB ; Germany}
\authorblockA{\tt{\small{$^1$firstname.lastname@ipvs.uni-stuttgart.de ~$^3$lastname@tu-berlin.de}}}
\vspace{-0.3cm}
}

\begin{document}

\maketitle
\thispagestyle{empty}
\pagestyle{empty}


\begin{abstract}

In this paper, we introduce a Grasp Manifold Estimator (GraspME) to detect grasp affordances for objects
directly in 2D camera images.
To perform manipulation tasks autonomously it is crucial for robots to have such \textit{graspability} models
of the surrounding objects. Grasp manifolds have the advantage of providing continuously infinitely many grasps, which is not the case when using other grasp representations such as predefined grasp points.
For instance, this property can be leveraged in motion optimization
to define goal sets as implicit surface constraints in the robot configuration space.
In this work, we restrict ourselves to the case of estimating possible end-effector positions directly from 2D camera images. To this extend, we define grasp manifolds via a set of keypoints and locate them in images using a Mask R-CNN \cite{he2017mask} backbone. Using learned features allows to generalize to different view angle, with potentially noisy images, and objects that were not part of the training set.
We rely on simulation data only and perform experiments on simple and complex objects, including unseen ones.
Our framework achieves an inference speed of 11.5 fps on a GPU, an average precision for keypoint estimation of 94.5\% and a mean pixel distance of only 1.29. This shows that we can estimate the objects very well via bounding boxes and segmentation masks as well as approximate the correct grasp manifold's keypoint coordinates.

\end{abstract}


\section{Introduction}

As humans share tasks with robots that are increasingly more autonomous,
it will become essential to provide user interfaces or robot behaviors that allow to
flexibly define the task objectives.
Hence in a human-robot collaborative manipulation task, knowledge of the entire object's grasp manifold
(i.e.  suitable grasp candidates), provides a step in this direction (e.g.  shared autonomy).
Note that the rapid detection of grasp manifolds in image space can have other applications
ranging from robot motion planning to character animation in video games.

\begin{figure}[t!]
  \centering
  \includegraphics[trim=0cm 4cm 2cm 2.5cm,clip,width=.9\columnwidth]{
  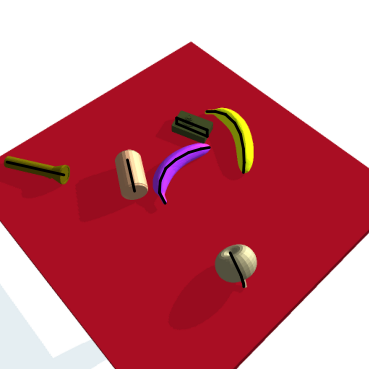}
  \caption{Predicted grasp manifolds 
   for complex objects using our approach, depicted as black lines.}
  \label{fig:title}
\end{figure}

In this paper, we a present a grasp manifold estimator \textit{GraspME}, based on the Detectron2 framework \cite{wu2019detectron2}.
Our model estimates the grasp manifolds, classifies the objects and computes their bounding boxes and segmentation masks all at the same time from a 2D image. A outcomes of such a grasp manifold estimation is depicted in Fig. \ref{fig:title}.
We train our model by supervised learning on simulation data
from an environment we develeopped in PyBullet \cite{coumans2019}.
We devised two sets of objects: the first with simple geometry and the second with more complex geometry from the 50 category subset of 3DNet \cite{wohlkinger20123dnet}.
Our simulation environment generates RGB, depth and segmentation images together with bounding boxes and grasp manifold keypoints for each object in a scene.

Object detection and semantic segmentation as well as grasp point localization have been improved steadily by the vision community which has led to a large list of baselines, e.g.  Mask R-CNN \cite{he2017mask} and several models based on its framework.
Solutions based on grasp point detection often either rely on predefined grasp points
or trial and error learning, which often proves to be rather unstable on unseen objects.
An important aspect is often neglected, namely that for most objects, infinitely many grasp points exist instead of just a few predefined.
This amount of grasp points can typically be defined by a manifold on a given object, mostly depending on the object geometry.

Thus, our contribution consists of 

\begin{enumerate}
    \item the introduction of the new problem setting of grasp manifold estimation on objects,
    \item \textit{GraspME}, a framework for object detection and grasp manifold keypoints estimation from 2D images and
    \item a simulation environment to generate suitable scenes and data for this task.
\end{enumerate}

Object's grasp manifold provide more knowledge about the scene
than simple grasp points. In human-robot collaboration this
means providing more solutions for a handover.
In space sharing scenario this grasp manifold
may lead to more reactive behavior fallingback
to different grasp solutions if the human moves
and thus minimally disrupting the human.

Another area that could benefit from grasp manifolds is Task and Motion Planning (TAMP). For example, it could be used in Logic Geometric Programming \cite{toussaint2015logic}, where optimization over continuously many grasp locations in a manifold - in contrast to fixing a specific grasp - could lead to a better trajectory with respect to the used control cost or even lead to more feasible and stable solutions of the problem.

This paper is structured as follows: In Section~\ref{sec:rel_work}, we present relevant related work. We then formulate the problem of grasp manifold estimation from images and
introduce notation in Section~\ref{sec:problem}.
Before introducing our GraspME framework and implementation
in Section~\ref{sec:method}, we present the dataset we work
with to train our estimator in Section~\ref{sec:data_generation}.
Finally, Sections~\ref{sec:experiments} and ~\ref{sec:conclusions} present our experiments, results and conclusions.
\vspace{.3cm}


\section{Related Work}
\label{sec:rel_work}

Our work merges the two research fields of object detection and grasp point detection. The first part is done by following Mask R-CNN \cite{he2017mask} while we combine it with a keypoint detection approach for the second.

\textbf{Mask R-CNN:} Mask R-CNN is a successor of Faster R-CNN \cite{ren2016faster} and Fast R-CNN \cite{girshick2015fast} and addresses the problem of object detection. It consists of two parts, a backbone to generate region proposals and a head to solve the actual task. To detect objects, Mask R-CNN's head is composed of two computation branches of which one is responsible for classifying the object and generating an axis aligned bounding box while the second branch estimates the object's segmentation mask. Mask R-CNN has also been used before to detect unseen objects in simulation in \cite{danielczuk2019segmenting}. Regarding keypoint estimation, Mask R-CNN suggested the extension of their framework with a third branch to estimate human poses using keypoints. Our own framework extends this approach by applying it on a new problem of predicting grasp manifolds and their corresponding keypoints.

\textbf{Grasp Point Detection:} A common approach to detect grasp points is to predefine fixed points on objects and localize them to plan a grasp trajectory, e.g.  \cite{spenrath2017gripping}. These methods often lack in generalizability, as they cannot work properly on unseen objects without grasp point labels. Another possibility is to let the model learn to differentiate between good and bad grasp points and extrapolate the knowledge to unseen objects. To be able to do this distinction, the model can either rely on predefined grasp points or more common by executing random grasps on the objects and learn by trial-and-error \cite{levine2018learning, mahler2017dex, morrison2018closing, morrison2020learning}. 

A benchmark for object affordances has been recently introduced \cite{deng20213d} to evaluate point cloud deep learning networks. Our framework complements the aforementioned approaches by introducing manifolds that contain possible grasp points from which an algorithm can sample and execute a grasp.

\textbf{Keypoint Detection:} The topic of keypoint detection is often associated to Human Pose Estimation. There are also several datasets for this problem, e.g.  MPII \cite{andriluka20142d} or COCO \cite{lin2014microsoft}. Several approaches address Human Pose Estimation via keypoint detection and incorporating domain knowledge, e.g.  the COCO  2016 keypoint detection winner \cite{cao2017realtime}.

However, keypoints detection has lately also being used to tackle other problems such as improving the quality of image generation \cite{he2021latentkeypointgan}
or object detection by estimating the object's center as keypoint \cite{zhou2019objects}. We use keypoints to describe the grasp manifolds and estimate them during inference.


\begin{figure}[b]
	\centering
	\includegraphics[width=0.32\columnwidth]{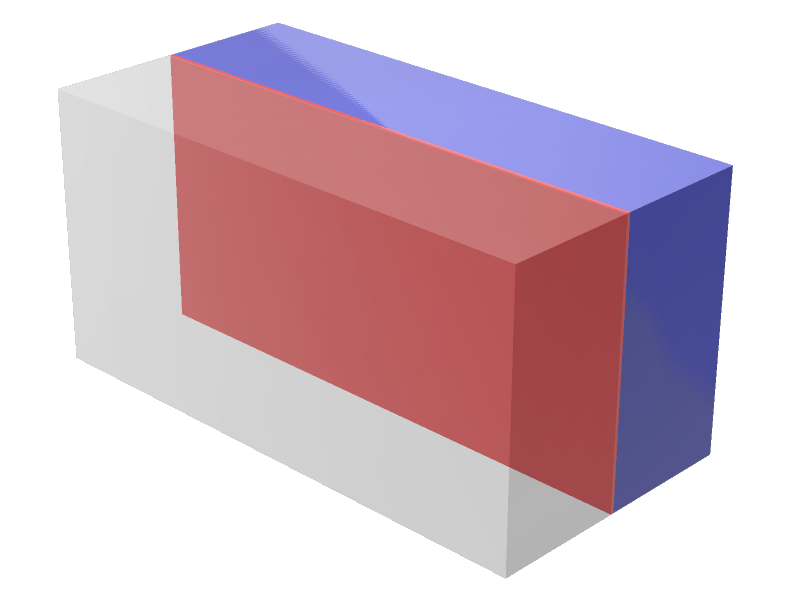}
	\includegraphics[width=0.32\columnwidth]{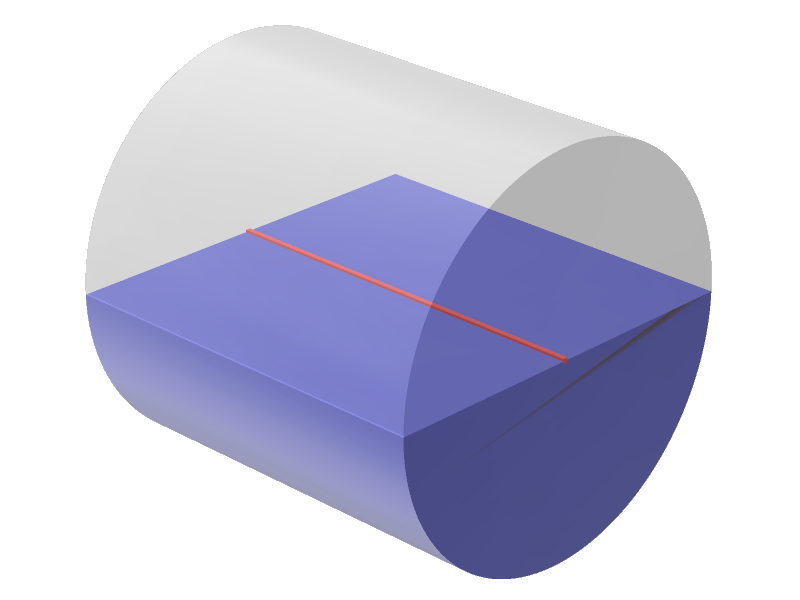}
	\includegraphics[width=0.32\columnwidth]{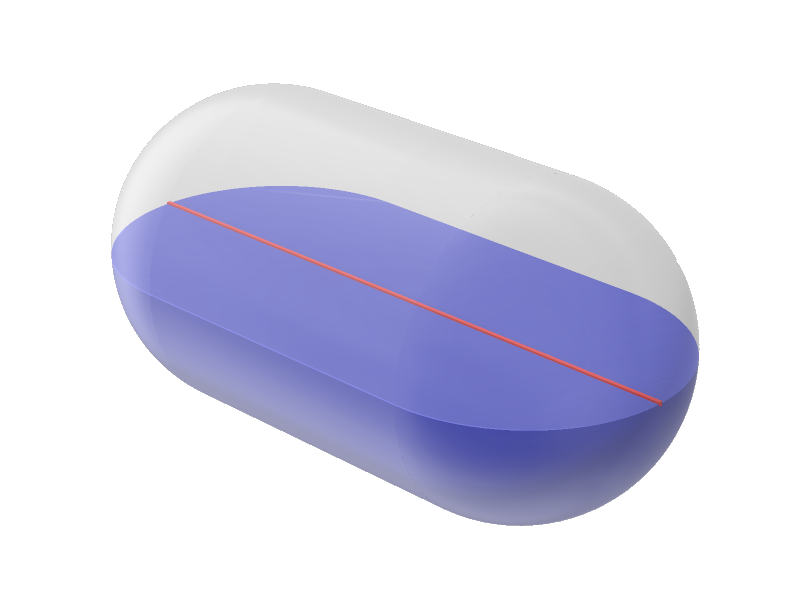}
	\caption{The three object models in blue and transparent with the corresponding grasp manifold in red: cuboid, cylinder, capsule (left to right).}
	\label{fig:objects_simple}
\end{figure}

\section{Grasp Manifolds}
\label{sec:problem}


\subsection{Definition}

We formalize the concept of an object's grasp manifold as a region $GM$ with a continuous closed border. Any point $p \in GM$ defines a potential grasp on the object, such that the closing point of the gripper is the same as this point $p$. 

To simplify the problem, we approximate such a grasp manifold by a set of keypoints, to which we will refer to as \textit{grasp manifold keypoints} $kp_i$. Typically, they are chosen as corner points that span the corresponding manifold, thus approximating it as close as possible.

\subsection{Object and Gripper Geometry}

Depending on the object's geometry, the corresponding grasp manifold can be defined as a line or a whole surface (see next subsections).

We assume the usage of a parallel two finger gripper, i.e.  to perform a grasp using an object's grasp manifold, the gripper should be aligned parallel to the manifold such that the closing point of the gripper would be on the chosen grasp point on the manifold.
We expect that this can easily be extended to other types of gripper.


\subsection{Simple Objects}


We assume that the simple objects have a lengthy shape, i.e.  one of the sides along the axes is longer than the other, i.e.  the object's \textit{main axis}. In most cases, the main axis defines already the grasp manifold, e.g. for cylinders and capsules. If we neglect any other situational circumstances, it is possible to perform a grasp on a cylinder or a capsule if the gripper is aligned parallel to the main axis of the object.


Therefore, the grasp manifold for capsules and cylinders is defined by the starting and the ending keypoint of the object's main axis going through the object's center, resulting in a line.
An example of the grasp manifold for these two object types can be seen in Fig. \ref{fig:objects_simple} as a red line.

For the cuboids, the manifold containing possible grasp points is much larger. The main axis of the cuboid is still the key to define it. However, it can expand in the upper and lower direction as well, creating a plane parallel to the cuboid's faces, as can be seen on the left in Fig. \ref{fig:objects_simple} in red.
For simplicity, we reduced the grasp manifold for cuboids to a line (the main axis), such that only two keypoints define the grasp manifold for simple objects.


\subsection{Complex Objects}

For objects with less trivial geometries we defined the grasp manifold manually by specifying the corresponding keypoints. To simplify the problem, we restrict ourselves to a maximum of $K$ keypoints per object which approximates the real grasp manifold.


The approximated grasp manifold is defined by connecting the sequence of keypoints $\{kp_i\}_{i=1}^{k}$ with $k \in [2, K]$, conditioned by the number of keypoints used to approximate the grasp manifold for the corresponding object. The grasp manifold of objects like bananas or bottles is a line, while it is a surface for objects like cameras and guitars.

\begin{figure}[t!]
	\centering
	\includegraphics[width=0.9\columnwidth]{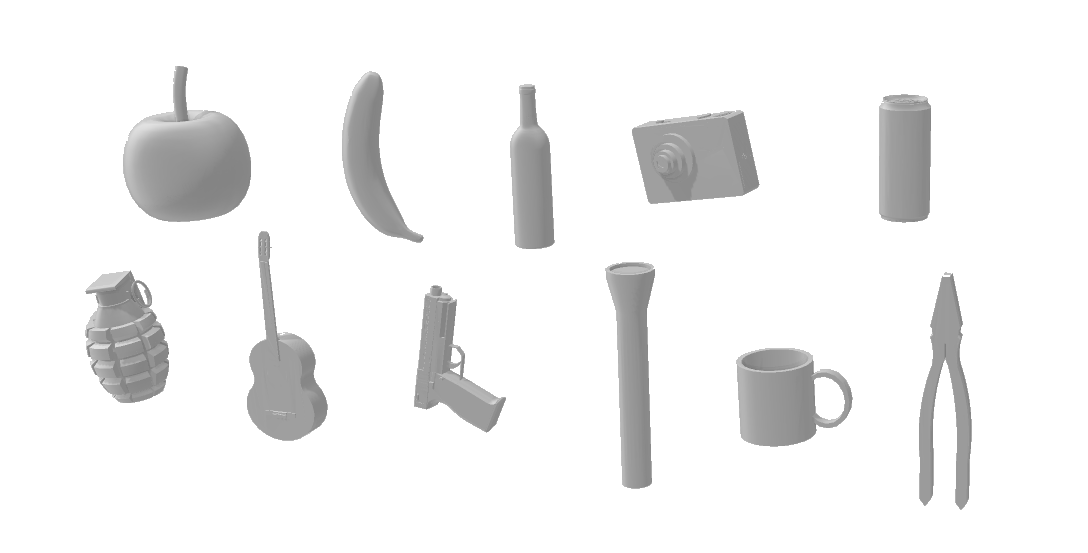}
	\caption{The complex object models taken from 3DNet \cite{wohlkinger20123dnet}.}
	\label{fig:objects_complex}
	\vspace{-.5cm}
\end{figure}


\section{The GraspME Model}
\label{sec:method}

Our proposed model estimates the grasp manifold from 2D images via keypoints localization instead of directly computing the object's pose, as mentioned in \ref{sec:problem}.


Since this procedure is analog to human pose estimation, we use the method of Mask R-CNN \cite{he2017mask} and the corresponding Detectron2 framework \cite{wu2019detectron2} as basis for our model.


\subsection{Architecture}

\begin{figure}[t!]
	\centering
	\includegraphics[width=0.95\columnwidth]{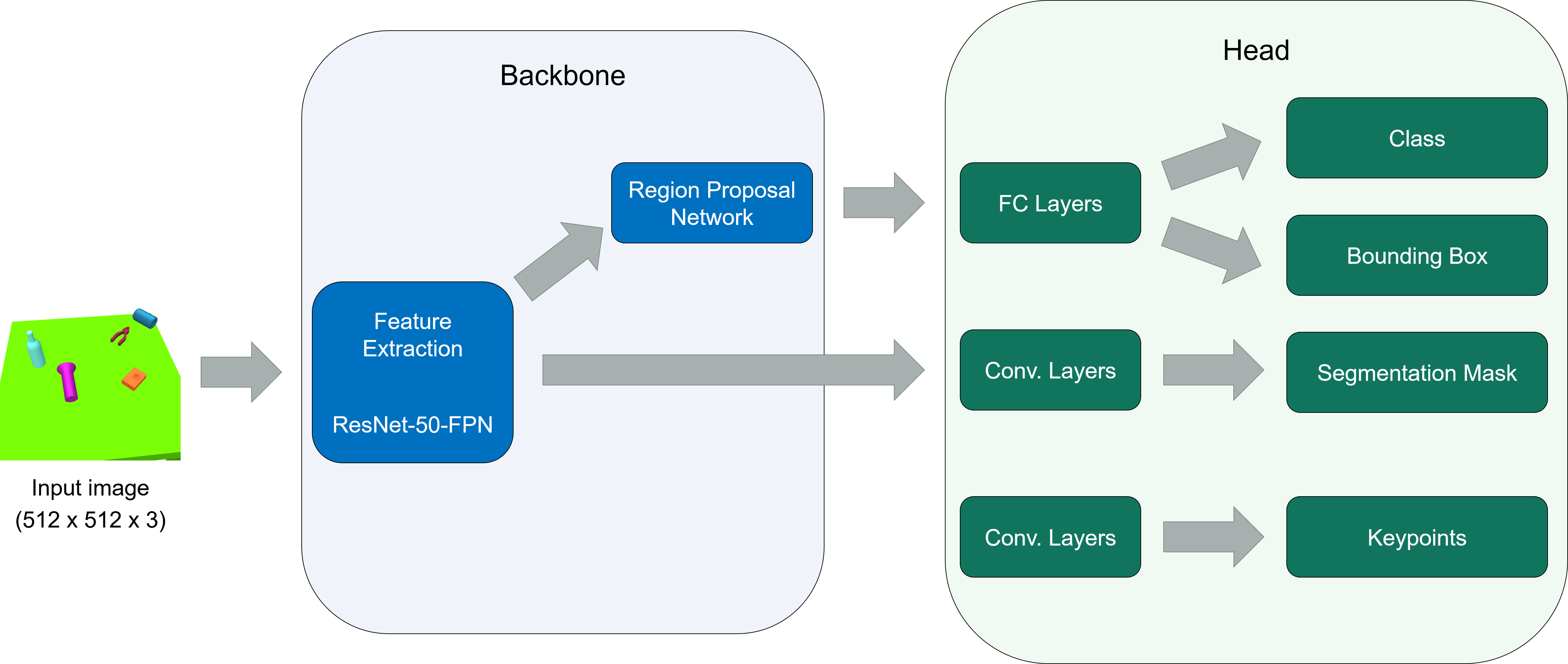}
	\caption{GraspME framework for grasp manifold estimation.}
	\label{fig:architecture}
\end{figure}

Similar to Mask R-CNN, our framework also consists of two main parts.
First, a backbone model, the Region Proposal Network (RPN) from Faster R-CNN \cite{ren2016faster}, is used on the whole image to generate region proposals. These are fed into the network's region of interest (ROI) head for the actual task: the classification, the bounding box detection, the mask prediction and the keypoints estimation. The overall framework architecture can be seen in Fig. \ref{fig:architecture}.


For the backbone, we use the ResNet-FPN variant from Mask R-CNN \cite{he2017mask}. It consists of a ResNet \cite{he2016deep} of depth 50, denoted as ResNet-50, or of depth 101, denoted as ResNet-101, with a Feature Pyramid Network (FPN) \cite{lin2017feature} on top. The rest of the architecture follows the suggestions from Mask R-CNN and the Detectron2 framework \cite{wu2019detectron2} for the Human Pose Estimation. For our experiments, we use only the 2D RGB images as input.

Depending on the object types, we set the number of keypoints to detect to $K = 2$ for the simple objects and to $K = 10$ for the complex objects. We chose this number because we found that we do not need more than 10 to approximate manifolds for our objects. Extending to more keypoints would be possible. Since not every complex object type needs 10 keypoints to define the corresponding grasp manifold, we add extra keypoints at the object's origin with a visibility flag equal to 0 such that each object type has a set of 10 keypoints and these additional keypoints will be neglected during the training.


Since our approach should be able to detect unknown objects, all object types belong to the same category ``object'' which leads to a class-agnostic object detection task. However, we additionally train a network with different object classes for comparison.

\section{Training Data Generation}
\label{sec:data_generation}



Our simulation environment is based on PyBullet \cite{coumans2019}. The generated data should be diverse enough such that the trained model generalizes to unseen poses and objects, and to real data.
	
We consider the scenario where a robot grasps objects from a flat table surface while observing the scene from above. The camera's position is sampled from a hemisphere around the tabletop's center. The camera records RGB and depth data together with corresponding segmentation images of the observed scenes. The segmentation images are automatically generated by the PyBullet simulation.

Additionally, we store the axis aligned bounding boxes per object and the keypoints that define the grasp manifold for the corresponding object. 

\subsection{Bounding Boxes}
The bounding boxes are computed by using the minimal and maximal x- and y-coordinates of the object's segmentation mask. This way, we define the bounding box with the lower point $(x_{min}, y_{min})$, its width $w = x_{max} - x_{min}$ and its height $h = y_{max} - y_{min}$. Due to occlusions, the segmentation mask and the bounding box are computed only for the object's visible part.

\subsection{Keypoints}



We define the keypoints for each object type beforehand, and project them during the simulation on the image plane using the full projection matrix of the camera, giving us the absolute position of a grasp manifold keypoint $kp_i = (x_i, y_i)$.


Since a keypoint could be invisible due to being occluded or outside of the camera's view, we set a visibility flag $v_i$ for them using the COCO format \cite{lin2014microsoft} for keypoint detection, i.e.  $v_i = 0$ if the keypoint does not exist on the object, $v_i = 1$ if the keypoint exists but is not visible and $v_i = 2$ if the keypoint is visible. Furthermore, for the simple objects only, if $kp_1$ is not visible but $kp_2$ is, we swap $kp_1$ and $kp_2$ and their corresponding visibility flags such that $kp_1$ should always be visible. This is possible due to the symmetric properties of those objects.

\subsection{Object Shapes and Randomization}

\begin{figure*}[t!]
  \centering
  \begin{subfigure}{\columnwidth}
      \includegraphics[width=0.24\columnwidth]{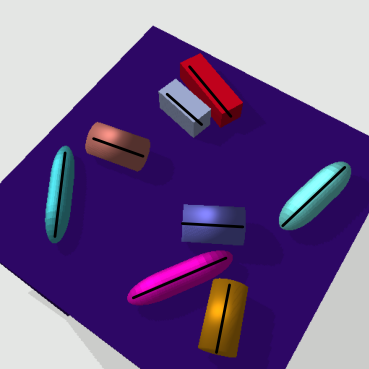}
      \includegraphics[width=0.24\columnwidth]{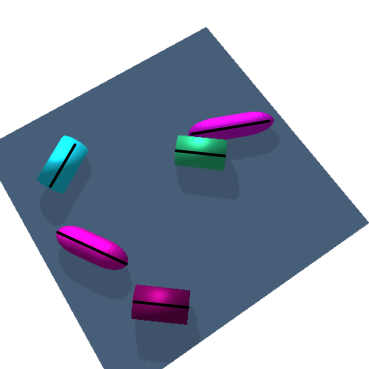}
      \includegraphics[width=0.24\columnwidth]{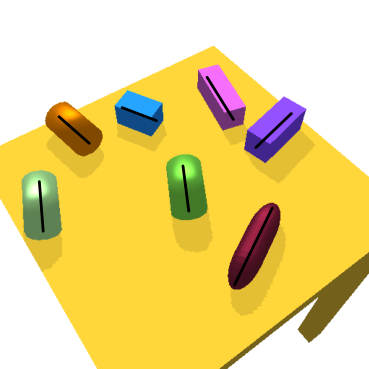}
      \includegraphics[width=0.24\columnwidth]{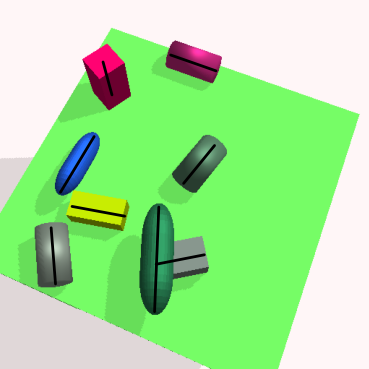}
      \caption{Ground-truth for simple objects}
      \label{fig:gt_sim}
  \end{subfigure}
  \begin{subfigure}{\columnwidth}
      \includegraphics[width=0.24\columnwidth]{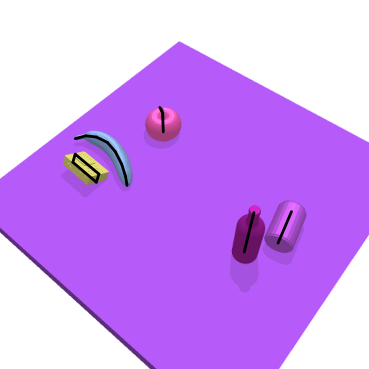}
      \includegraphics[width=0.24\columnwidth]{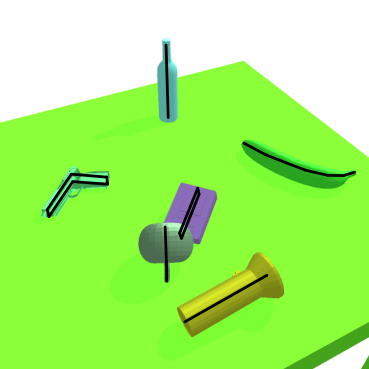}
      \includegraphics[width=0.24\columnwidth]{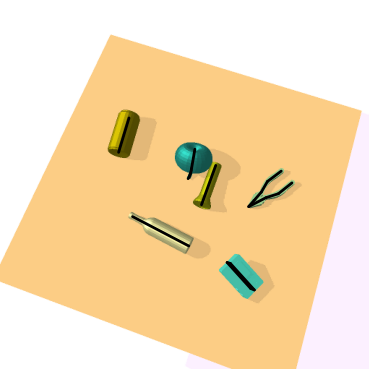}
      \includegraphics[width=0.24\columnwidth]{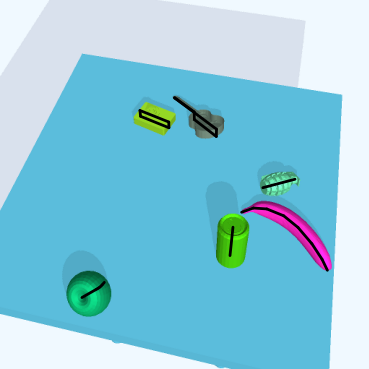}
      \caption{Ground-truth for complex objects}
      \label{fig:gt_com}
  \end{subfigure}\\
  \vspace{0.01\columnwidth}
  \begin{subfigure}{\columnwidth}
      \includegraphics[width=0.24\columnwidth]{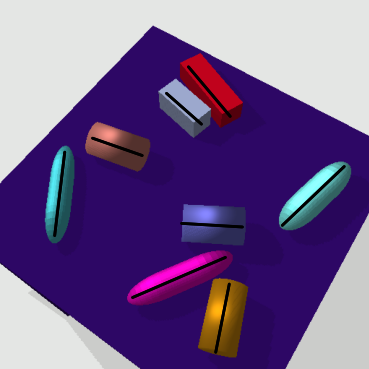}
      \includegraphics[width=0.24\columnwidth]{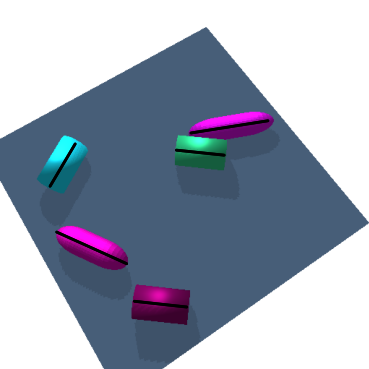}
      \includegraphics[width=0.24\columnwidth]{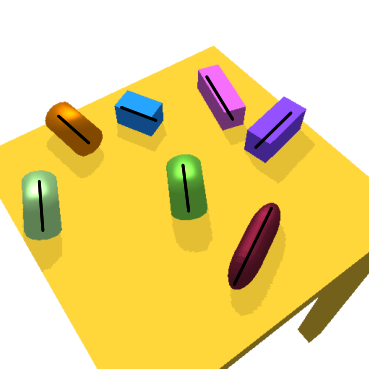}
      \includegraphics[width=0.24\columnwidth]{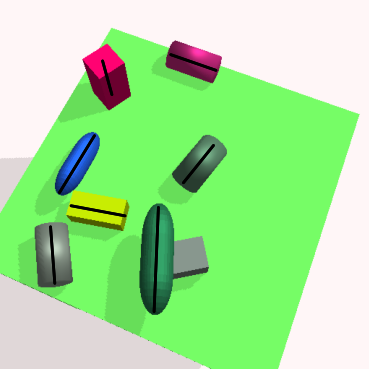}
      \caption{Predictions for simple objects on R101-S-M-80}
      \label{fig:pred_sim}
  \end{subfigure}
  \begin{subfigure}{\columnwidth}
      \includegraphics[width=0.24\columnwidth]{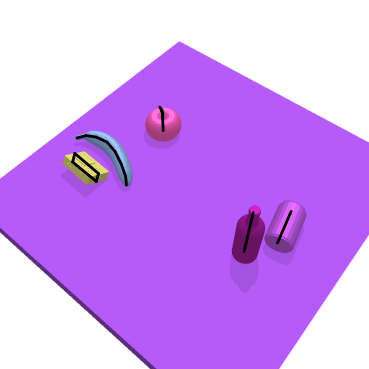}
      \includegraphics[width=0.24\columnwidth]{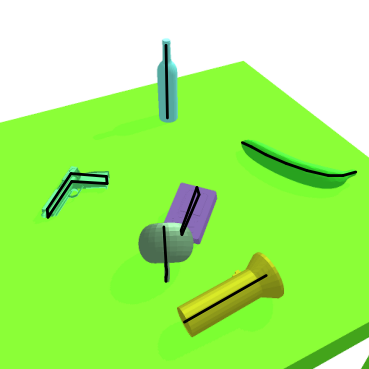}
      \includegraphics[width=0.24\columnwidth]{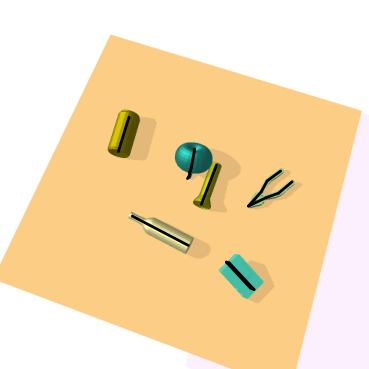}
      \includegraphics[width=0.24\columnwidth]{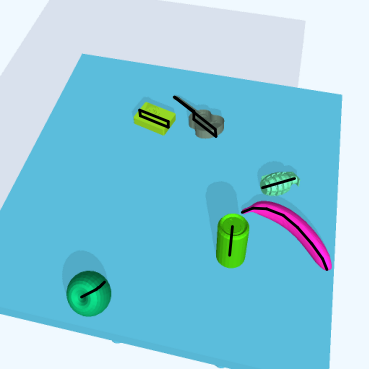}
      \caption{Predictions for complex objects on R101-C-M-40}
      \label{fig:pred_com}
  \end{subfigure}
  \caption{Results of the predictions (bottom) in comparison to the ground-truth (top) for simple (left) and complex objects (right). The corresponding grasp manifolds are depicted as black lines.}
  \label{fig:results_simcom}
\end{figure*}

For cuboids, cylinders and capsules, the sizes are randomized for each object instance before rendering them, i.e.  the cuboid's length, width and height are chosen such that the length has the largest value with a small probability of generating cubes. The same holds for the capsule's and the cylinder's length and radius. Typical samples during the simulation can be seen in Fig. \ref{fig:gt_sim}.

For complex objects, we use a small subset of 11 objects from the 50 category subset of 3DNet \cite{wohlkinger20123dnet}: apple, banana, bottle, camera, can, grenade, guitar, gun, maglite, mug and pliers. These synthetic 3D mesh models are rescaled and transformed from the original versions to fit our simulation. The object models are depicted in Fig. \ref{fig:objects_complex} while some samples from the simulation with complex objects can be seen in Fig. \ref{fig:title} and \ref{fig:gt_com}.

To increase the diversity of the simulated dataset, we incorporate domain randomization techniques which have been proposed by Tobin et al. \cite{tobin2017domain}.
Hence, we randomize for each scene separately, the camera's view point, the lighting conditions with one light source, the table's color, the total amount of objects and the objects' colors, positions and orientations before dropping them from above 
the tabletop using simulated physics.

\begin{table}[t]
\caption{ An overview over the training parameters chosen for the different architectures and experiments. The model name is a concatenation of abbreviations of the training parameters in the following order: backbone (R50/R101), training data (S/C/P), class agnostic (merged classes = M) or classification (C), iterations (40/80)}
\label{tab:train_parameters}
\begin{center}
\tiny
\begin{tabular}{|c||c|c|c|c|c|}
\hline
Model Name & Backbone & Class Agnostic & Training Data & Test Data & Iterations\\
\hline
\hline
R50-S-M-40 & \multirow{3}{*}{ResNet-50-FPN} & True & \multirow{5}{*}{Simple} & \multirow{5}{*}{Simple} & 40k\\
\cline{1-1} \cline{3-3} \cline{6-6}
R50-S-C-40 &  & False &  &  & 40k\\
\cline{1-1} \cline{3-3} \cline{6-6}
R50-S-M-80 &  & True &  &  & 80k\\
\cline{1-3} \cline{6-6}
R101-S-M-40 & \multirow{2}{*}{ResNet-101-FPN} & True &  &  & 40k\\
\cline{1-1} \cline{3-3} \cline{6-6}
R101-S-M-80 &  & True &  &  & 80k\\
\cline{1-6}
R50-C-M-40 & \multirow{3}{*}{ResNet-50-FPN} & True & \multirow{5}{*}{Complex} & \multirow{5}{*}{Complex} & 40k\\
\cline{1-1} \cline{3-3} \cline{6-6}
R50-C-C-40 &  & False &  &  & 40k\\
\cline{1-1} \cline{3-3} \cline{6-6}
R50-C-M-80 &  & True &  &  & 80k\\
\cline{1-3} \cline{6-6}
R101-C-M-40 & \multirow{2}{*}{ResNet-101-FPN} & True &  &  & 40k\\
\cline{1-1} \cline{3-3} \cline{6-6}
R101-C-M-80 &  & True &  &  & 80k\\
\cline{1-6}
\multirow{3}{*}{R50-P-M-40} & \multirow{3}{*}{ResNet-50-FPN} & \multirow{3}{*}{True} & \multirow{3}{*}{Part} & Part & \multirow{3}{*}{40k}\\
\cline{5-5}
 &  &  &  & Unseen & \\
\cline{5-5}
 &  &  &  & Complex & \\
\cline{1-6}
\multirow{3}{*}{R50-P-M-80} & \multirow{3}{*}{ResNet-50-FPN} & \multirow{3}{*}{True} & \multirow{3}{*}{Part} & Part & \multirow{3}{*}{80k}\\
\cline{5-5}
 &  &  &  & Unseen & \\
\cline{5-5}
 &  &  &  & Complex & \\
\cline{1-6}
\end{tabular}
\end{center}
\end{table}

\subsection{Training}

The training procedure for the RPN, the classification, the bounding box detection, the mask prediction and the keypoint estimation are adopted from the Mask R-CNN which is why we refer readers for further details to \cite{he2017mask} and its predecessors Fast R-CNN \cite{girshick2015fast} and Faster R-CNN \cite{ren2016faster}.

During training, we apply some randomly chosen online augmentations to improve the generalizability of our model. We use the implemented augmentations from the Detectron2 framework \cite{wu2019detectron2} like flipping and changes of lighting, saturation, brightness and contrast, from which 2 augmentations are chosen at random for each batch.

\paragraph{Dataset}
We generated 40,000 synthetic scenes per object set. The model is then trained on 80\% of the data, i.e.  on 32,000 data points, while the remaining data is withheld for validation and testing, each containing 4,000 data points.

The images are of size $512 \times 512$ pixels which we keep fixed as input for our models. We collect these amounts of data for three datasets: the first one contains only simple objects (called ``Simple''), the second one contains only complex objects (called ``Complex'') and the third contains 8 out of 11 complex objects (called ``Part''). A fourth dataset including 4,000 data points each for validation and for testing contains the remaining three objects that did not appear in any scene in ``Part'' (called ``Unseen'').

\paragraph{Hyper parameters}

We chose an image batch size of 10 while the batch size per image is 64 for the RPN and 128 for the ROI head due to memory restrictions. As we cannot compare our method to any baseline due to lacking one, we use different training parameters as follows to get a better overview over the effects of these parameters.

The models are trained for 40k and 80k iterations with a base learning rate of 0.001. We decrease the learning rate by a factor of 10 after 30k iterations if trained for 40k iterations and additionally after 60k if trained for 80k iterations. Due to the similar approach, we use the two pretrained models from Mask R-CNN for the Human Pose Estimation experiments as initialization and finetune them on our datasets and problem setting. The training of the models is performed on two GeForce GTX 1080 Ti GPU. An overview over all model configurations can be found in Table \ref{tab:train_parameters}.


\begin{table}[t]
	\caption{ The results on the simple and complex test data w.r.t. AP in percent (\%), IoU of the grasp manifold in percent (\%) and mean pixel distance between keypoints on the three main tasks of bounding box, mask and keypoint estimation.}
	\label{tab:experiments_simcom}
	\begin{center}
		\tiny
		\begin{tabular}{|c||c|c|c||c|c|c|}
			\hline
			Model Name & AP$^{bb}$ & AP$^{seg}$ & AP$^{kp}$ & IoU$_{clip}$ & IoU$_{full}$ & mDist\\
			\hline
			\hline
			Random & - & - & 0.0 & 0.6 $\pm$ 1.5 & 0.6 $\pm$ 1.5 & 38.84 $\pm$ 8.01\\
			\hline
			R50-S-M-40 & 93.3 & 93.2 & 22.7 & 39.7 $\pm$ 32.8 & 39.7 $\pm$ 32.8 & 9.18 $\pm$ 9.93\\
			\hline
			R50-S-C-40 & 93.8 & 93.6 & 24.4 & 39.8 $\pm$ 32.4 & 39.8 $\pm$ 32.4 & 8.82 $\pm$ 9.44\\
			\hline
			R50-S-M-80 & 93.3 & 93.5 & 23.5 & 39.9 $\pm$ 32.5 & 39.9 $\pm$ 32.5 & 8.33 $\pm$ 9.35\\
			\hline
			R101-S-M-40 & \textbf{94.2} & \textbf{94.1} & 23.0 & 39.2 $\pm$ 32.8 & 39.2 $\pm$ 32.8 & 9.33 $\pm$ 10.19\\
			\hline
			R101-S-M-80 & 93.9 & 94.0 & \textbf{24.6} & \textbf{40.2} $\pm$ 32.5 & \textbf{40.2} $\pm$ 32.5 & \textbf{8.12} $\pm$ 9.90\\
			\hline
			\hline
			Random & - & - & 0.0 & 8.2 $\pm$ 10.9 & 10.1 $\pm$ 11.0 & 26.61 $\pm$ 8.68\\
			\hline
			R50-C-M-40 & 96.4 & 93.1 & 93.5 & 64.8 $\pm$ 29.6 & 26.1 $\pm$ 22.2 & 1.52 $\pm$ 2.79\\
			\hline
			R50-C-C-40 & 94.9 & 91.2 & 91.5 & 64.4 $\pm$ 29.7 & \textbf{28.4} $\pm$ 24.1 & 1.57 $\pm$ 3.09\\
			\hline
			R50-C-M-80 & 96.5 & 93.1 & 93.4 & 64.8 $\pm$ 29.5 & 25.3 $\pm$ 21.5 & 1.51 $\pm$ 2.85\\
			\hline
			R101-C-M-40 & \textbf{96.7} & \textbf{93.4} & \textbf{93.7} & 65.2 $\pm$ 29.8 & 25.8 $\pm$ 25.2 & 1.53 $\pm$ 2.92\\
			\hline
			R101-C-M-80 & 96.5 & 93.2 & 93.2 & \textbf{65.4} $\pm$ 29.8 & 25.6 $\pm$ 25.0 & \textbf{1.48} $\pm$ 2.87\\
			\hline
		\end{tabular}
	\end{center}
\end{table}

\section{Experiments}
\label{sec:experiments}


The experiments are conducted on the two scenarios with simple objects and with complex objects. Since there does not exist any baseline yet that estimates whole grasp manifolds from 2D images, we can only compare different architectures and training setups of our models. An overview of the performed experiments for all of the model's outputs can be seen in Table \ref{tab:train_parameters}. Overall, we achieve an average speed of 11.5 frames per second on one of the before mentioned GPUs which makes our framework suitable for real time applications.

\subsection{Metrics}

We evaluate our models using the standard COCO metrics \cite{lin2014microsoft} regarding average precision (AP) for the three outputs bounding box (bb), segmentation mask (segm) and keypoints (kp). As the metric for the evaluation of the latter output is optimized for human pose estimation, we additionally compute the mean Intersection over Union (IoU) of the ground-truth grasp manifold with the predicted one by using the same number of keypoints as the ground-truth (clip) or the full set of predicted keypoints (full) as well as its standard deviation.

We also measure the mean pixel distance (mDist) of the predicted keypoints to the ground-truth ones by matching them first with the closest ground-truth keypoint set per object. Afterwards, we conduct some ablation studies on the different model architectures and training setups of our models. Due to a lacking baseline, we compare our model to the Random baseline, i.e.  we randomly sample keypoints from the predicted bounding boxes from the models R50-S-M-40 for the ``Simple" test dataset and R50-C-M-40 for the ``Complex'', ``Part'' and ``Unseen'' test datasets.

\begin{table}[t]
	\caption{ The results on the complex test data including unseen objects w.r.t. average precision (AP) in percent (\%) on the three main tasks of bounding box, mask and keypoint estimation.}
	\label{tab:experiments_part}
	\begin{center}
		\tiny
		\begin{tabular}{|c|c||c|c|c||c|c|c|}
			\hline
			Data & Iter. & AP$^{bb}$ & AP$^{seg}$ & AP$^{kp}$ & IoU$_{clip}$ & IoU$_{full}$ & mDist\\
			\hline
			\hline
			\multirow{3}{*}{Part} & Rand. & - & - & 0.0 & 9.6 $\pm$ 12.1 & 12.0 $\pm$ 12.1 & 26.34 $\pm$ 9.07\\
			\cline{2-8}
			& 40k & \textbf{96.6} & \textbf{92.4} & \textbf{94.5} & 68.7 $\pm$ 31.2 & \textbf{30.5} $\pm$ 23.4 & 1.30 $\pm$ 3.04\\
			\cline{2-8}
			& 80k & 96.5 & 92.3 & 94.3 & \textbf{69.1} $\pm$ 31.1 & 28.9 $\pm$ 21.8 & \textbf{1.29} $\pm$ 2.96\\
			\hline
			\multirow{3}{*}{Unseen} &  Rand. & - & - & 0.0 & 5.0 $\pm$ 6.4 & 6.0 $\pm$ 6.0 & 27.76 $\pm$ 7.99\\
			\cline{2-8}
			& 40k & \textbf{60.9} & \textbf{86.0} & \textbf{1.9} & \textbf{13.6} $\pm$ 15.6 & \textbf{12.1} $\pm$ 9.7 & \textbf{21.09} $\pm$ 8.12\\
			\cline{2-8}
			& 80k & 57.4 & 84.9 & 1.8 & 13.4 $\pm$ 15.7 & 12.1 $\pm$ 9.8 & 21.43 $\pm$ 8.20 \\
			\hline
			\multirow{3}{*}{Comp.} &  Rand. & - & - & 0.0 & 0.6 $\pm$ 1.5 & 0.6 $\pm$ 1.5 & 38.84 $\pm$ 8.01\\
			\cline{2-8}
			& 40k & \textbf{89.4} & \textbf{92.2} & \textbf{56.4} & 51.0 $\pm$ 37.7 & \textbf{24.9} $\pm$ 22.1 & \textbf{7.77} $\pm$ 7.04\\
			\cline{2-8}
			& 80k & 88.0 & 91.8 & 54.1 & \textbf{51.1} $\pm$ 37.7 & 23.6 $\pm$ 20.6 & 7.88 $\pm$ 7.48\\
			\hline
		\end{tabular}
	\end{center}
\end{table}

\subsection{Simple Objects}


As can be seen from the upper part of Table \ref{tab:experiments_simcom}, we achieve a very high accuracy with all our models for the bounding box detection and the segmentation mask estimation while the AP for keypoints is rather low. 
\Janik{The reason for this is the objects' symmetry, which makes it harder to estimate the correct keypoints without any additional information. Flipping the keypoint labels during training could solve this problem, by specializing on certain image regions, i.e.  $kp_1$ could always be rather on the left of the object while $kp_2$ could always be on the right.}
	
As can be seen in Fig. \ref{fig:pred_sim}, if we detect an object, we indeed predict the keypoints and the corresponding grasp manifolds very well in comparison with its ground-truth in Fig. \ref{fig:gt_sim}. The models with the ResNet-101-FPN backbone achieve the highest accuracies as well as the best values for our own metrics.


We get values around 40\% for the grasp manifold IoUs, which proves that our assumptions about the low AP$^{kp}$ values is correct. This might also be the reason for the rather high mean pixel distances between the ground-truth and the predicted keypoints. As all of the objects had two ground-truth keypoints which were fully used, the values for IoU$_{clip}$ and IoU$_{full}$ are the same. Our model outperforms the Random baseline by far in all of the keypoint related metrics.

\subsection{Complex Objects}

The results of the bounding box detection and the segmentation mask estimation, reported in the lower part of Table \ref{tab:experiments_simcom}, are off similar quality as for simple objects. However, due to the unique shape of the objects, the AP$^{kp}$ values are much higher as the keypoints are much easier to identify on the objects.

This is also reflected in the low mean pixel distance between ground-truth and prediction of around 1.6 pixels which also leads to a grasp manifold IoU of 65\%. Though using the full set of available keypoints does not seem to be helpful as the IoU is much smaller. 

Objects with less keypoints do not benefit from the additional keypoints as their grasp manifold is approximated already well enough. Due to the larger number of keypoints, the Random baseline achieves higher scores with our metrics but is still a lot worse than our framework. Overall, the best results are again achieved by the models with the ResNet-101-FPN backbone. These accurate results can also be seen in Fig. \ref{fig:pred_com}, where we predict all of the keypoints nearly perfect in comparison with its ground-truth in Fig. \ref{fig:gt_com}.

\begin{figure}[t!]
	\centering
	\includegraphics[width=0.24\columnwidth]{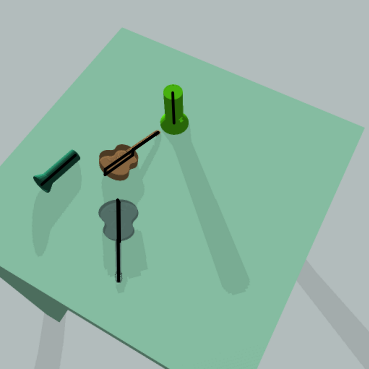}
	\includegraphics[width=0.24\columnwidth]{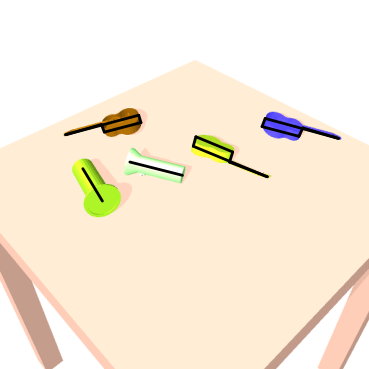}
	\includegraphics[width=0.24\columnwidth]{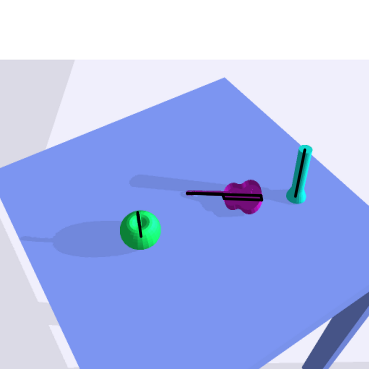}
	\includegraphics[width=0.24\columnwidth]{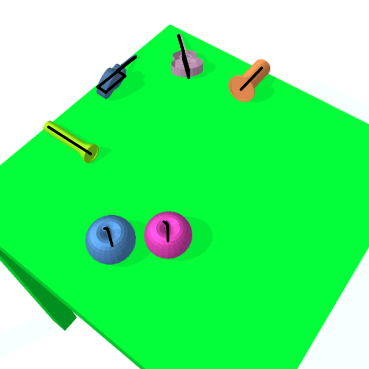}\\
	\vspace{0.01\columnwidth}
	\includegraphics[width=0.24\columnwidth]{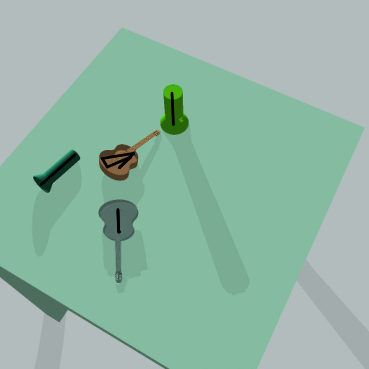}
	\includegraphics[width=0.24\columnwidth]{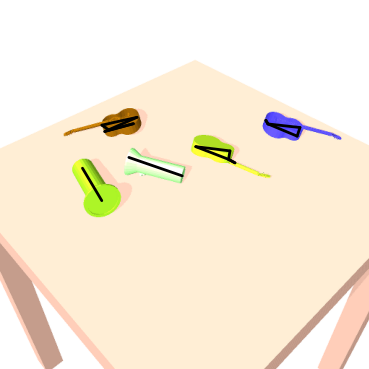}
	\includegraphics[width=0.24\columnwidth]{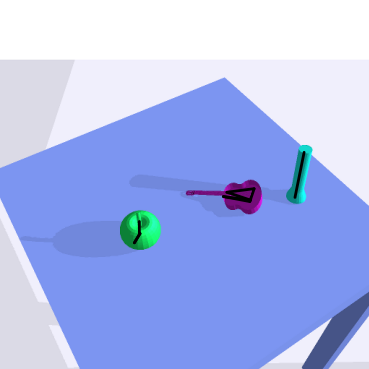}
	\includegraphics[width=0.24\columnwidth]{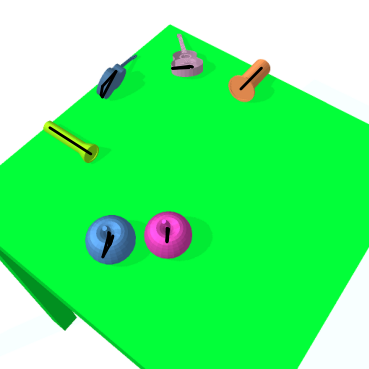}
	\caption{Results of the predictions (bottom) in comparison to the ground-truth (top) for unseen objects on R50-P-M-40. The corresponding grasp manifolds are depicted as black lines.}
	\label{fig:results_unseen}
\end{figure}

\subsection{Unseen Objects}

For the third experiment, we want to evaluate the generalizability of our models by using the dataset ``Part'' to train on a subset of the objects and test on the dataset ``Unseen'' containing some withheld objects. Additionally, we provide results on the full dataset ``Complex'' with all objects. We report the corresponding overview in Table \ref{tab:experiments_part}.

As can be seen, the models trained and evaluated on the ``Part'' dataset achieve similar performance to the models trained on the ``Complex'' dataset with respect to the COCO metrics while getting even better scores with our own metrics in terms of mean IoU with nearly 70\% and mean pixel distance to the ground-truth keypoints of around 1.3 pixels. This could be due to the withheld objects that might belong to the objects that are more difficult.

Even though the model has never encountered objects from the ``Unseen'' dataset, it could still segment most of them from the images and estimate corresponding bounding boxes. However, computing the expected keypoints from the ground-truth was not possible, following the COCO metrics and the high pixel distance. There is some intersection of the grasp manifolds though, as can be seen from the IoU values, from which we can assume that a grasp manifold has still been found by the models.

Using the full amount of keypoints decreases the quality of the results. We assume that these low values overall come from the very different object shapes in comparison to the known object's shapes and that it was difficult for the model to approximate the expected grasp manifold and the corresponding keypoints. However, the model might have predicted another unintentional grasp manifold that is still a valid grasp manifold. By extending our approach to find several grasp manifolds or using more than one ground-truth grasp manifold per object, we might be able to achieve better results.

To emphasize this hypothesis, we compare some of the results on the unseen objects in Fig. \ref{fig:results_unseen}. As can be seen, the result's quality depends on the object. The grasp manifolds for the maglites are quite accurate while the grasp manifolds for the apples also seem to be rather close to the ground-truth. As the guitar is the most difficult of these objects regarding shape, the results are not so good in comparison with the ground-truth. However, the model still often predicts a grasp manifold by focusing on the guitar's corpus which seems to be valid, even if it is different from the expected one. We conclude that our framework is able to estimate grasp manifolds also on unseen objects.

As the ``Complex'' dataset contains both seen and unseen objects, the COCO scores are obviously lower as for the models trained on all objects but we still achieve very good results in all three categories and also estimating the grasp manifolds well enough as can be seen from the IoU. Therefore, having some unseen objects mixed with known objects does not decrease the results too much, except for the mean pixel distance. The model might even benefit from having these mixed scenes and hence, achieve better results. Even for the ``Unseen'' dataset as well as for the ``Part'' and the ``Complex'' datasets, our framework outperforms the Random baseline.

\begin{figure}[t!]
  \centering
  \includegraphics[width=0.24\columnwidth]{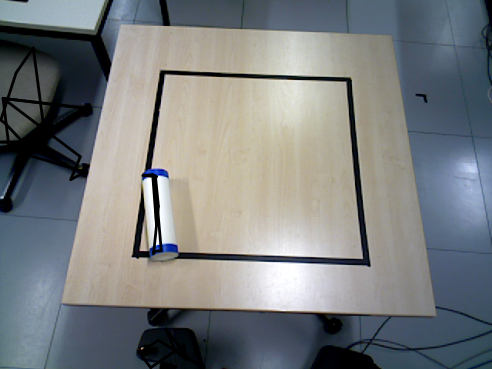}
  \includegraphics[width=0.24\columnwidth]{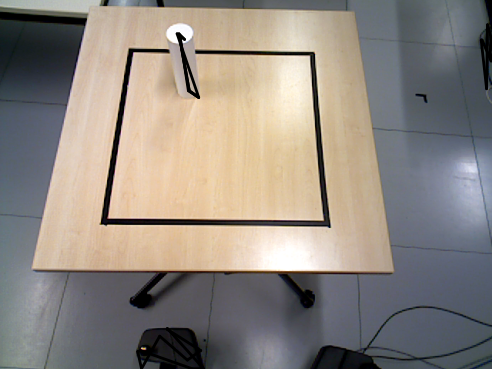}
  \includegraphics[width=0.24\columnwidth]{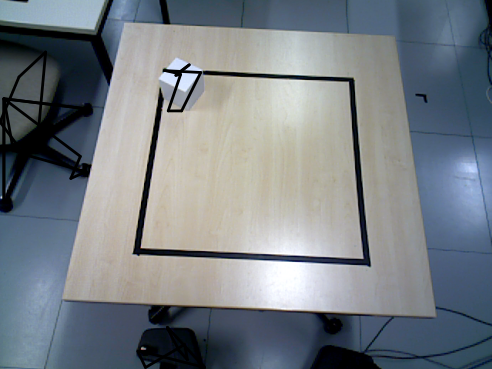}
  \includegraphics[width=0.24\columnwidth]{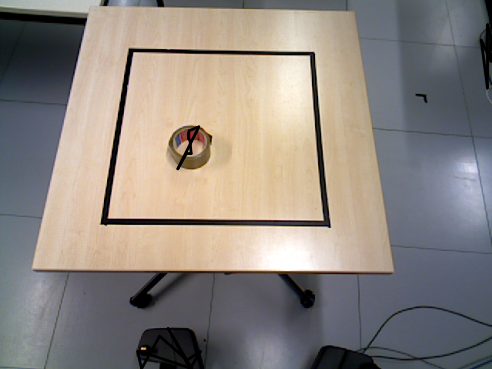}
  \caption{Results of the predictions for unseen real data on R101-C-M-40. The corresponding grasp manifolds are depicted as black lines.}
  \label{fig:results_real}
\end{figure}


\subsection{Ablation Studies}

To further evaluate our framework, we compare different model architectures and training setups as ablation studies. The first part is about the model's architecture in terms of the backbone. Regarding the results of our experiments, we can see that models with the ResNet-101-FPN backbone achieved better results than with the ResNet-50-FPN backbone.

Training the model by additional 40k iterations does not seem to improve the results as much as expected regarding the COCO metrics. For the ResNet-101-FPN backbone, the model trained for only 40k instead of 80k iterations achieves even better results. The higher number of training iterations is only  noticeable with respect to the IoU and the mDist.

The difference between using separate classes for each object type or simply having one for all, i.e.  having the class-agnostic case, is rather unsignificant. For the complex objects, the class-agnostic model achieves better results by nearly 2 points in all three categories of the COCO metrics and slightly better values for the IoU and the mDist metrics. Therefore, we recommend the class-agnostic model as it can also be used for unseen objects without additional training.

\subsection{Real Data}

We present initial results on real camera images. However, as ground truth labels were not available, we could only evaluate the resulting images. Some of the better results are depicted in Fig. \ref{fig:results_real}, which seems promising regarding the usage in real applications, as in most cases the object's main axis is predicted.


\section{Conclusion}
\label{sec:conclusions}

Overall, we showed that our models achieve good results in terms of object detection for both simple and complex objects. Even though the values for evaluating the models on unseen objects are low, we can see that our framework could still partially generalize to these shapes and predict a grasp manifold. Thus, our model can support other methods for finding suitable grasp points on objects by spanning a whole manifold of possibilities. By having a frame rate of 11.5 fps, we expect to use our approach for real time applications.

We plan to provide a proof of concept by integrating it into a trajectory optimization framework and demonstrate our model's usage to perform human-robot-collaboration tasks, e.g. via shared autonomy and autonomous environment interactions in real scenarios. The sampled grasps in these scenarios will also be compared to those provided by other algorithms mentioned in Section~\ref{sec:rel_work}. Furthermore, we want to extend our framework in taking advantage of additional depth data to gain valuable information about the scene.

As this problem setting is unknown yet, we hope to draw interest to this scenario and encourage other researchers to approach it and use our framework as baseline.


\section*{Acknowledgment}

We want to thank the authors of \cite{he2017mask} and \cite{wu2019detectron2} for making the code of their framework publicly available.
This   work   was   conducted   while   Ruben Bauer was  performing his Masters dissertation in the Machine Learning and Robotics Lab, University of Stuttgart, Germany.
This work is partially funded by the research alliance
``System Mensch''.



\newpage

\bibliography{bibliography}
\bibliographystyle{ieeetr}



\end{document}